\documentclass[a4paper]{article}

\usepackage[english]{babel}
\usepackage[utf8]{inputenc}
\usepackage[T1]{fontenc}

\usepackage{listings}

\usepackage{lmodern}
\usepackage[autostyle]{csquotes}

\usepackage[a4paper,top=3cm,bottom=2cm,left=3cm,right=3cm,marginparwidth=1.75cm]{geometry}

\usepackage{amsmath}
\usepackage{graphicx}
\usepackage[colorinlistoftodos]{todonotes}
\usepackage{authblk}
\usepackage{rotating}
\usepackage{csvsimple}
\usepackage{float}

\usepackage{multicol}

\usepackage[colorlinks=true, allcolors=blue]{hyperref}
\usepackage{rotating}

\title{A Field Guide to Scientific XAI: Transparent and Interpretable Deep Learning for Bioinformatics Research}
\author[1*]{Thomas P. Quinn}
\author[1]{Sunil Gupta}
\author[1]{Svetha Venkatesh}
\author[1]{Vuong Le}

\affil[1]{\footnotesize Applied Artificial Intelligence Institute (A2I2), Deakin University, Geelong, Australia

* \textit{contacttomquinn@gmail.com}
}
\date{}

\Affilfont{\fontsize{4}{4}}

\begin{document}
\maketitle

\begin{abstract}
Deep learning has become popular because of its potential to achieve high accuracy in prediction tasks. However, accuracy is not always the only goal of statistical modelling, especially for models developed as part of scientific research. Rather, many scientific models are developed to facilitate scientific discovery, by which we mean to abstract a human-understandable representation of the natural world. Unfortunately, the opacity of deep neural networks limit their role in scientific discovery, creating a new demand for models that are transparently interpretable. This article is a field guide to transparent model design. It provides a taxonomy of transparent model design concepts, a practical workflow for putting design concepts into practice, and a general template for reporting design choices. We hope this field guide will help researchers more effectively design transparently interpretable models, and thus enable them to use deep learning for scientific discovery.
\end{abstract}

\section{Introduction to Transparency}

The recent success of deep learning in text and image processing has ushered in many new scientific applications. In the field of bioinformatics, deep learning is already being used to predict clinical phenotypes \cite{duvallet_meta-analysis_2017}, DNA-binding motifs \cite{koo_deep_2020}, RNA splicing sites \cite{jaganathan_predicting_2019}, and much more \cite{eraslan_deep_2019}. However, the most empirically accurate among these models are notoriously ``black-box'', meaning that the inner workings of the model are opaque to the human user. This \textit{opacity} is intrinsic to the model itself, and arises from a mismatch between the mathematical operations of a deep neural network and the nature of human-scale reasoning \cite{burrell_how_2016}.

Although deep models are potentially highly accurate, accuracy is not always the only goal. Scientific models are often used for scientific discovery, by which we mean to abstract a human-understandable representation of the natural world.
Unfortunately, the opacity of deep neural networks limit their role in scientific discovery, creating a new demand for models that \textit{explain} to the user how they work (and hence are \textit{explainable}). This demand is not unique to bioinformatics \cite{lipton_mythos_2017}, and already there exists many innovative ways to make machine learning models more understandable to humans. In fact, \textit{explainable artificial intelligence} (XAI) has become a field in its own right \cite{barredo_arrieta_explainable_2020}, with conceptual advances that could prove useful for bioinformatics applications \cite{azodi_opening_2020}, notably with regard to scientific discovery. We use the term \textit{scientific XAI} to refer to the use of XAI for scientific discovery. Unlike XAI, which only aims to describe how a model behaves, scientific XAI aims to describe how the underlying natural world behaves via a description of how a model behaves, typically using data generated by a carefully designed experiment.

There are two general approaches to scientific XAI. The first aims to use a standalone algorithm to explain the behaviour of any opaque model (called the \textit{model-agnostic explainability} approach). The second aims to use a custom-built model purposefully designed to have transparency (called the \textit{model-specific transparency} approach). Taking the inverse of the opacity definition, we define \textit{transparency} as an intrinsic property of a model whose mathematical operations align with the nature of human-scale reasoning. While there is no universal definition of interpretable \cite{lipton_mythos_2017}, we will use a contextual definition, and consider a transparent model to be interpretable if an intended user can, in a given context, obtain a desired interpretation from the model.%

Although both approaches have merit, this article focuses on model-specific transparency, which we believe lends itself nicely to many scientific XAI applications (and yet has received less attention in the literature).
In the following sections, we
(1) discuss the pros and cons of using transparent models;
(2) provide a taxonomy of transparent model design concepts;
(3) offer a practical workflow for putting design concepts into practice; and
(4) propose a general template for reporting design choices. 
We hope this field guide will help researchers more effectively design transparently interpretable models, and thus enable them to use deep learning for scientific discovery.

\section{Pros and Cons of Transparent Models}

In the pursuit of scientific discovery, an analyst must first choose between \textit{model-agnostic explainability} or \textit{model-specific transparency}. This is not always an easy decision to make, and (like many things) depends on the research question being asked. Instead of offering prescriptive advice, we present arguments for and against each approach.

On one hand, model-specific transparency is purpose-built for an application, making all interpretations intrinsic to the model itself. For example, a line is a kind of interpretable model where the coefficients can be inspected directly as a measure of how much each independent variable influences the dependent variable. This confers a certain kind of reliability that model-agnostic explainability lacks, or put colloquially, ``what you see is what you get''.
However, transparent models require more time and skill to implement. Depending on the application, the design of a transparent model may be a research project on its own.

\begin{figure}[H]
\centering
\scalebox{1}{
\includegraphics[width=(.8\textwidth)]{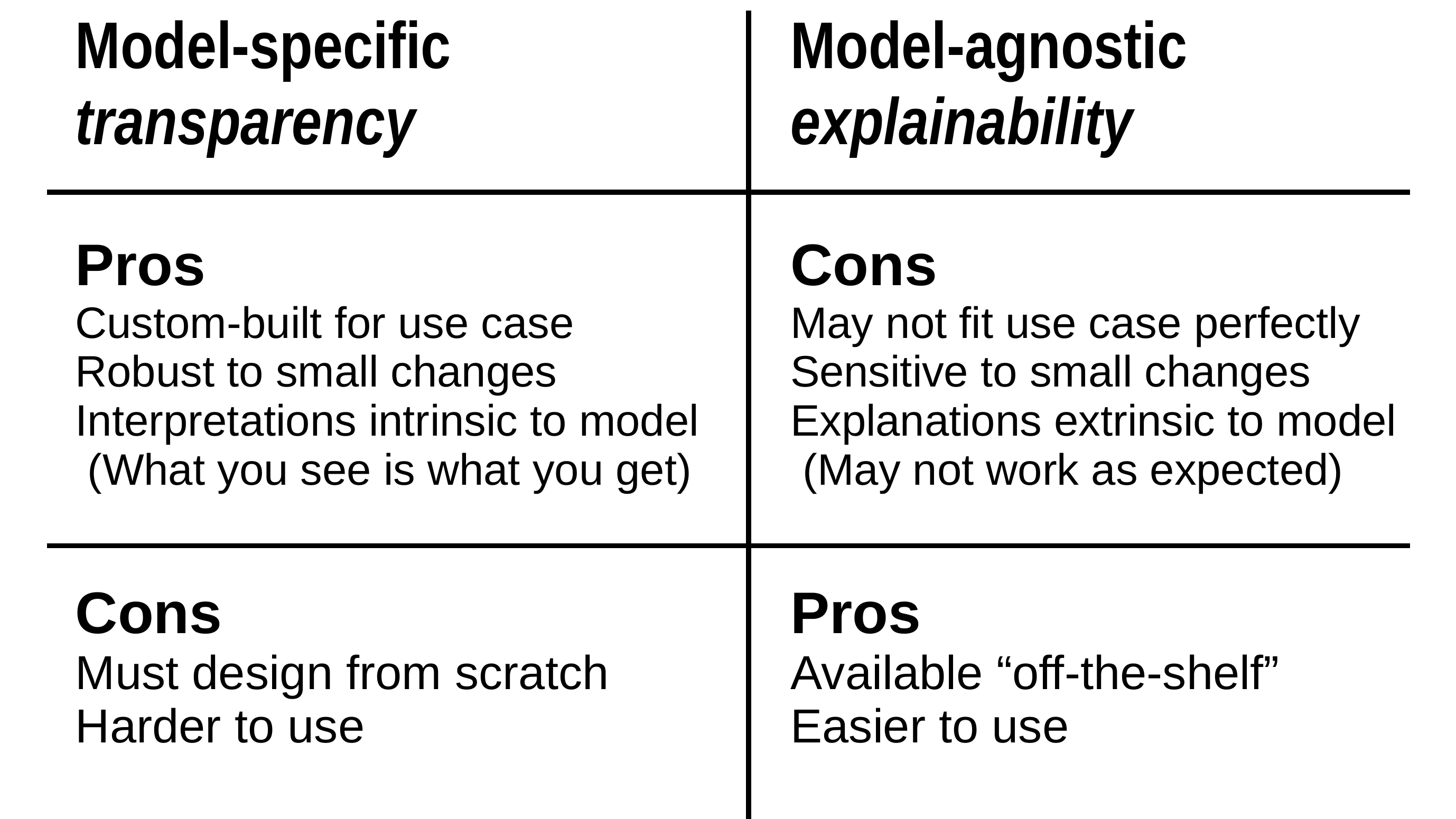}}
\caption{The pros and cons to consider when choosing between \textit{model-agnostic explainability} vs. \textit{model-specific transparency} for scientific discovery. This article focuses on model-specific transparency, in particular the design of transparent models.}
\label{fig:pros-cons}
\end{figure}

On the other hand, model-agnostic explainability is convenient to use because it de-couples the task of model design from the task of model interpretation. As a first step, the analyst can train an arbitrarily complex deep neural network to achieve good performance. As a second step, the analyst can try out a number of model-agnostic methods to explain the behavior of their model. Alternatively, one might skip the first step and instead explain a model previously created by someone else. For example, a complex neural network can be approximated by a surrogate linear model whose coefficients are inspected directly in place of the neural network weights \cite{ribeiro_why_2016}.
However, explanations produced by model-agnostic methods can be sensitive to small changes in the data, raising concerns that these methods may not work as expected \cite{alvarez-melis_robustness_2018}. Rudin argued against model-agnostic explanations for high-stakes applications, noting that model-agnostic explanations must be inadequate, because otherwise only the model explanations, and not the underlying model itself, would be needed in the first place \cite{rudin_stop_2019}. Figure~\ref{fig:pros-cons} summarizes these pros and cons.

For readers who want to use model-agnostic explainability, we refer them to the second half of the Interpretable Machine Learning textbook by Christoph Molnar \cite{molnar_interpretable_2019}. The remainder of this article will deal with model-specific transparency, where we present a formal analysis of transparent model design. Specifically, we focus on the design of neural networks, especially those where the network weights are inspected directly to facilitate the interpretation, akin to how one would typically interpret linear models. In the next section, we lay out a taxonomy of general transparent model design concepts, which readers can think of as the Lego blocks of transparent model design. In the following section, we will move on to provide a workflow of transparent model design, which readers can think of as a general guide to Lego block assembly.

\section{Taxonomy of Transparency Methods}

The success of deep learning models comes from their ability to pass massive amounts of data through deeply nested layers of non-linear operations. The generally accepted rule-of-thumb is: \textit{the deeper the layers and the bigger the data, the better the model}. This may be true in terms of performance, but increasing the complexity of a model--as well as increasing its indiscriminate use of mass data--also tends to make the model more opaque, thus making it harder to use the model for scientific discovery.

\begin{figure}[H]
\centering
\scalebox{1}{
\includegraphics[width=(.8\textwidth)]{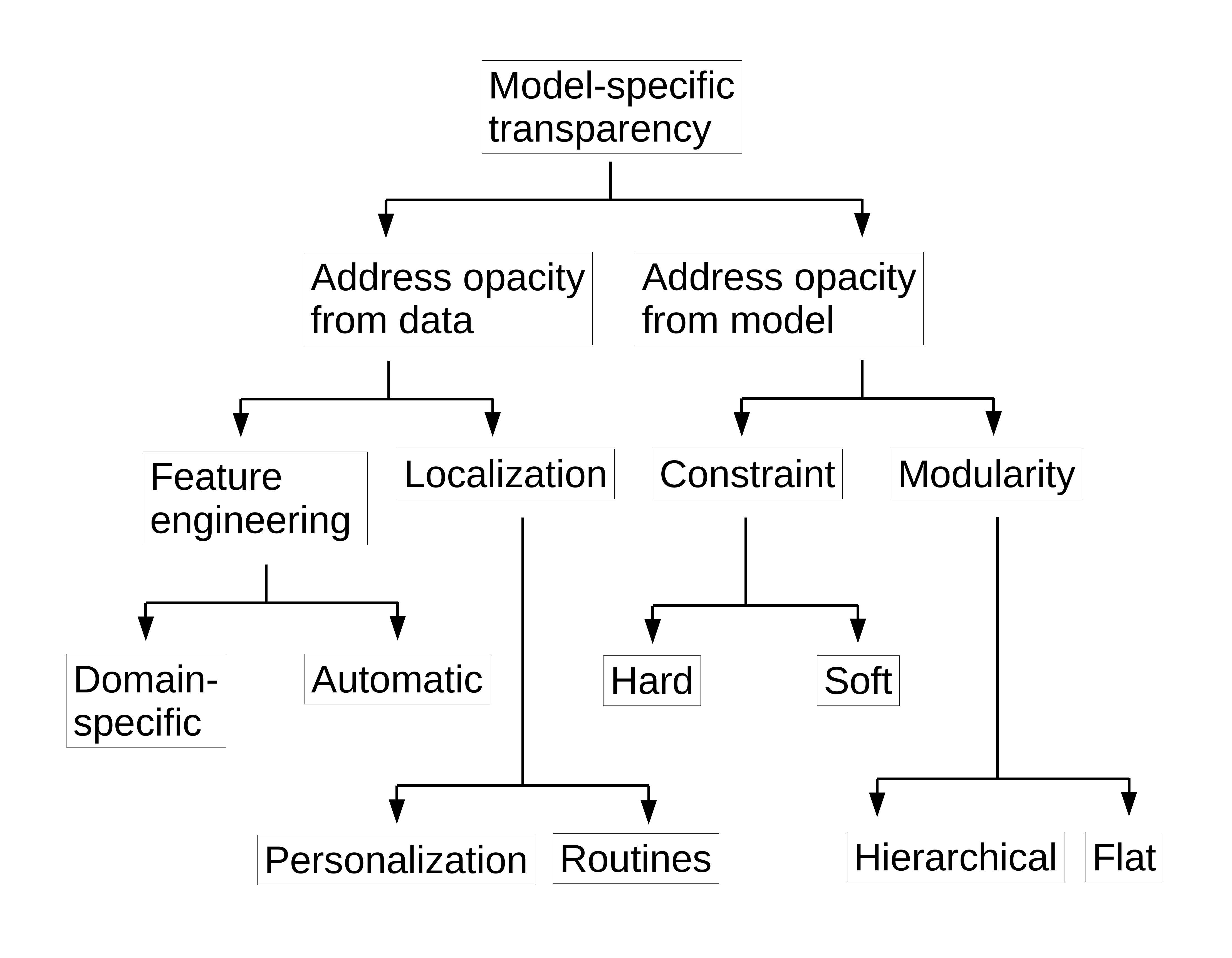}}
\caption{A taxonomy of transparent model design concepts.}
\label{fig:taxonomy}
\end{figure}

To implement transparency, the model must be designed from the start to resolve the opacity that arises from excess model complexity and indiscriminate data use. In other words, model design and model interpretation are intimately linked. Transparent model design is tailored to the problem under study, and so each new problem requires a new model. There are no hard and fast rules to transparent model design, but rather a collection of concepts that can be used to resolve the underlying sources of opacity. These concepts are used with the prior intention to build models that are \textit{intrinsically interpretable}, and hence are said to be \textit{a priori} interpretable (a term used to contrast this approach with \textit{post-hoc} interpretation).
We group these concepts into 4 categories: (1) \textbf{feature engineering}, (2) \textbf{localization}, (3) \textbf{constraint}, and (4) \textbf{modularity}. The first two address the opacity from the incoming data, while the latter two address the opacity from the model itself.
It is common, if not necessary, to combine several model-specific approaches at once. Figure~\ref{fig:taxonomy} shows a broad taxonomic tree of these approaches.

\subsection{Feature engineering}

Much of model interpretation involves interpreting the relationship between predictors (called \textit{features}) and outcomes. As such, a first consideration in transparent model design is the nature of the features themselves. A model is arguably more interpretable if the features are more interpretable. The synthesis of new features is called \textbf{feature engineering}, and is one of the most powerful tools in the modeller's toolbox. The goal of feature engineering is to introduce new structure into the data, typically by converting \textit{generic, unorganized signals} into \textit{specific, organized signals} so that the new signals--now structured as aggregates, sets, sequences, or graphs--more closely align with domain knowledge or human reasoning.
There are 2 general approaches:
\begin{itemize}
    \item \textbf{Domain knowledge-driven engineering} makes use of prior domain knowledge to synthesize new features. For example, one could convert gene expression signatures into a functional pathway score by adding up the expression levels of all genes belonging to each pathway \cite{beykikhoshk_deeptriage_2020}. As another example, one could convert DNA input into a table that describes a sequence of trinucleotide codons \cite{albaradei_splice2deep_2020}. Such engineering is typically done before model training.
    \item \textbf{Automatic engineering} takes a data-driven approach to synthesize new features without prior domain knowledge, most notably through an unsupervised or supervised machine learning method. In the unsupervised case, one could, for example, use non-negative matrix factorization to reduce high-dimensional data set into independent factors that describe different aspects of the data \cite{pauca_text_2004}. It is possible to interpret these factors by assigning them thematic labels based on the raw features they contain, as done in natural language processing \cite{pauca_text_2004}. In the supervised case, one could, for example, use one data set to learn a lower-dimensional representation of the feature input, then apply this representation to another data set (c.f., transfer learning \cite{zhuang_comprehensive_2020}). Automatic engineering can be done before model training. However, it can also be done during model training, for example by having a neural network layer learn a useful lower-dimensional representation of the data as part of end-to-end learning \cite{lakshmanan_machine_2020}. Either way, the resultant lower-dimensional representations are essentially newly engineered features. 
\end{itemize}
As a rule-of-thumb, models using engineered features are easier to train. For this reason, prior featuring engineering has an additional benefit beyond interpretability: they can also help models achieve higher performances, especially on smaller data sets where there may not be enough samples to learn useful lower-dimensional representations as part of end-to-end learning \cite{saeys_review_2007}.

\subsection{Localization}

\begin{figure}[H]
\centering
\scalebox{1}{
\includegraphics[width=(.8\textwidth)]{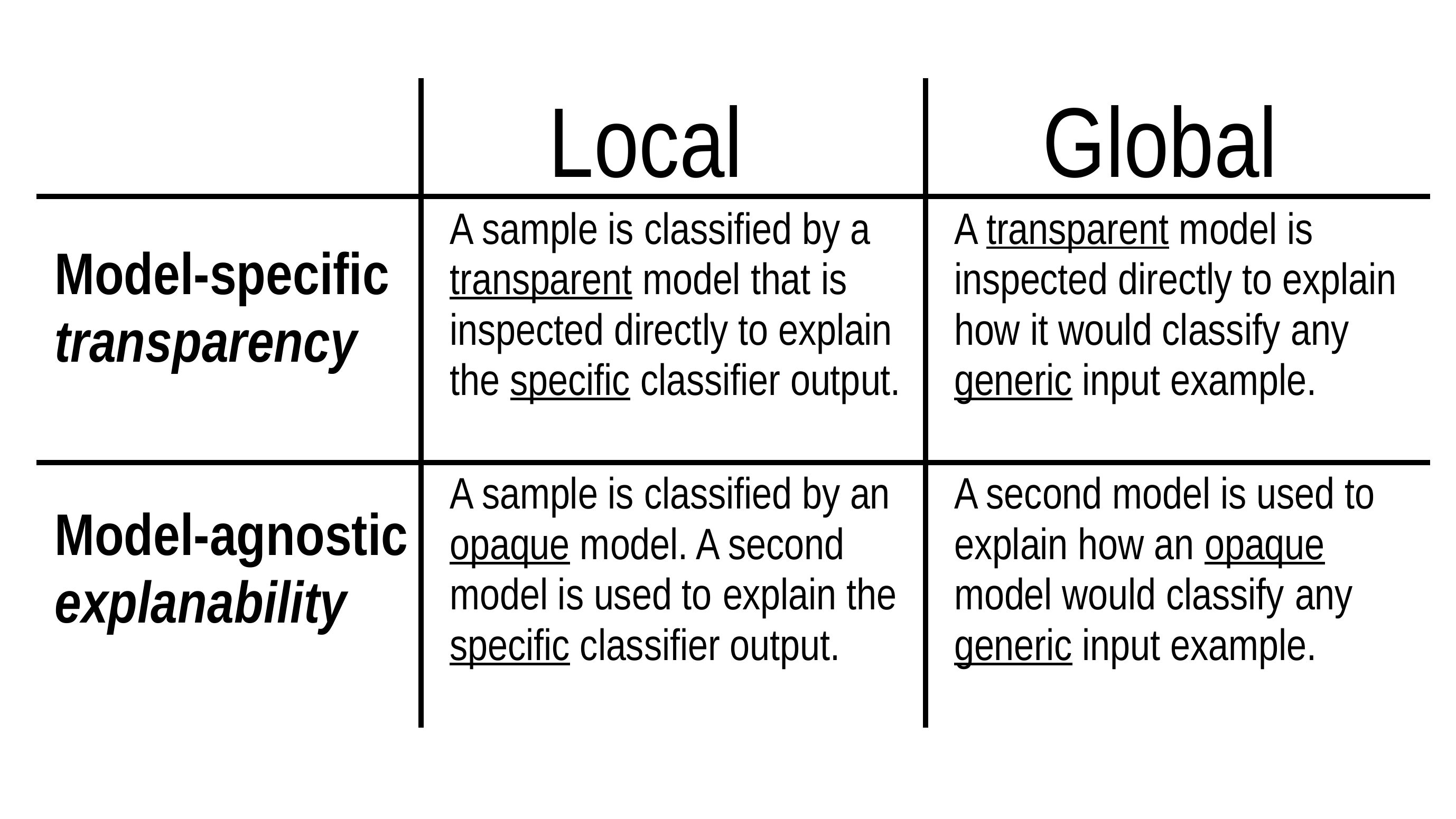}}
\caption{Types of model interpretations, using a classifier as an example. Descriptions of general behavior are called \textit{global}. Descriptions of specific behavior are called \textit{local}. Localization introduces local interpretations. Like transparency, explainability can also be made local.}
\label{fig:local-global}
\end{figure}

It is not always appropriate to interpret the features of data \textit{en masse}. In other words, some features may only be useful some of the time. This brings us to another major axis in model interpretability that 
delineates two distinct approaches to model interpretation. This axis is \textit{global} vs. \textit{local} interpretability. \textbf{Global} interpretations hold for the entire population of samples, while \textbf{local} interpretations hold for less than the entire population of samples. Figure~\ref{fig:local-global} further illustrates the two major axes of model interpretability, using a simple classifier as an example. While a global method may explain what features are important for any \textit{generic} sample's output, a local method can explain what features are important for a \textit{specific} sample's output. Such local interpretations operate under the assumption that what is important in one context may not be important in another context.

\textbf{Localization} introduces local interpretability into a model. One approach to localization is \textbf{personalization}, by which we mean reconfiguring a layer to produce sample-specific sets of parameters. Personalization is made possible through the attention mechanism \cite{bahdanau_neural_2016} or, similarly, self-explanation \cite{alvarez-melis_towards_2018}. Although attention is most often used to improve accuracy, it can be further leveraged to improve interpretability too: using attention, each sample can be explained by its own linear model, with coefficients that can be interpreted directly. Unlike the coefficients of a global linear fit that remain the same regardless of the sample considered, the coefficients of a local linear fit can be different for different samples.
Another approach to localization is sample-specific \textbf{routines} that use different processing channels for different types of data. For example, the fast-weights approach generates one set of parameters for one input type and another set for another input type \cite{ba2016using}. Applied to a biological problem, a single model designed to predict RNA splicing might consider different parameters depending on whether an acceptor site signature or a donor site signature is first detected in a sequence.

Note that local feature importance can emerge from statistical interactions within the data. For example, consider a simple regression $y = x + xz$. In a global sense, both $x$ and $z$ are important in the prediction of $y$. However, in a local sense, the importance of $z$ arguably depends on its context. For a sample where $x=0$, $z$ does not influence $y$, and so could be considered unimportant, at least for this sample. However, for a sample where $x$ is large, $z$ alone has a large influence on $y$, and so could be considered very important. Localization can help mine this kind of contextual importance \cite{quinn_deepcoda_2020}.
In any case, making sense of localization may require having meaningful features, and so localization can benefit from prior feature engineering.

\subsection{Constraint}

\textbf{Constraints} limit the complexity of a model, and enforce a logical or semantic meaning that allows the analyst to trace the relationship(s) between input and output. Constraints may be \textit{hard} in the sense of an explicit limit imposed by strict priors, or \textit{soft} in the sense of a simplification imposed by regularization terms. In practice, both hard and soft constraints may be used in combination. Generally speaking, constraints restrict the values or connectivity of the neural network weights, yielding a simpler model that may also be more interpretable.%
\footnote{By ``simpler'', we mean that the operations become more human-interpretable. Sometimes, a sophisticated architecture is needed to achieve this kind of simplicity. Building a simple model may not be a simple task!} We consider 4 constraints:

\begin{itemize}
    \item With \textbf{linearity}, an analyst preferentially uses linear (e.g., addition) or linear-like (e.g., ReLu activation) operations. For example, in generalized additive models (GAM), features may \textit{individually} undergo non-linear transformation, but are always combined linearly \cite{hastie_generalized_1986}.
    
    \item With \textbf{sparsity}, connections between neural network nodes are assigned zero value, and hence are disconnected. An architecture could limit the number of connections learned by the model (e.g., via an L1/L2-norm regularization as used by LASSO and elastic net), or could have the less relevant connections pruned after training \cite{lee_snip_2019}.
    
    \item With \textbf{non-negativity}, connections between neural network nodes are made non-negative (i.e., zero or positive). When combined with linearity, all operations become addition instead of subtraction. This may improve interpretability by reflecting how humans naturally reason by aggregating relevant signals together in an additive way \cite{chorowski_learning_2015}.
    
    \item With \textbf{discretization}, continuous values are replaced with discrete values, most often binary (i.e., zero or one). Related to this is the use of logical operations for machine learning, exemplified by the highly popular decision tree method. Although vanilla neural networks will typically only learn continuous values, continuous values can be made discrete through a secondary algorithm that rounds off weights to whole numbers \cite{gordon-rodriguez_learning_2021}.
\end{itemize}

\begin{figure}[H]
\centering
\scalebox{1}{
\includegraphics[width=(.6\textwidth)]{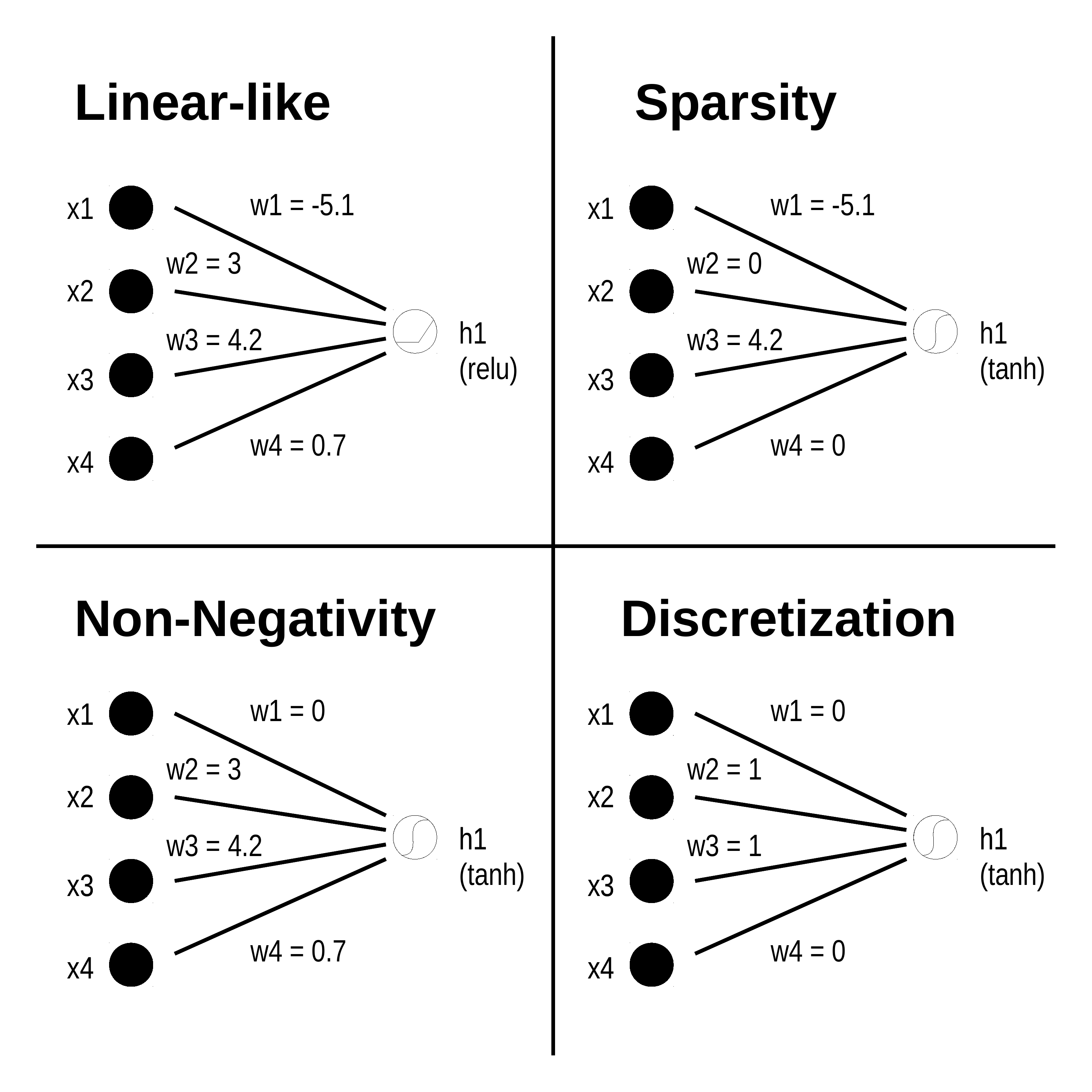}}
\caption{A visual depiction of 4 complexity constraints, showing the connections between a 4-node input layer and a single hidden node as an example. The 4 constraints are not mutually exclusive. Here, the discretized connections are also sparse and non-negative.}
\label{fig:constraints}
\end{figure}

These constraints, as depicted in Figure~\ref{fig:constraints}, are used primarily to limit the complexity of a model, and hence can be considered \textbf{complexity constraints}. Complexity constraints may be used for pragmatic reasons, for example to reduce the likelihood of over-fitting, or for theoretical reasons, for example to use a linear fit because the process being modelled is assumed to be additive. One popular complexity constraint is L1/L2-norm regularization, which can introduce sparsity by penalizing models with larger coefficients (a soft constraint because the model can still use larger coefficients if they sufficiently reduce the loss). Another example is GA(2)M, which extends GAM to account for multiplicative interactions between features, while still combining all terms linearly \cite{caruana_intelligible_2015} (a hard constraint because the model is restricted to have a pre-defined architecture).

Instead of a complexity constraint, one could use prior knowledge to introduce a \textbf{knowledge constraint}, whereby the architecture or regularization forces the model interpretations to align with a body of knowledge. One example of a knowledge constraint is \textbf{credibility}, which uses a regularization term to force the model to prefer, all things being equal, features labelled as important by experts \cite{wang_learning_2018} (a soft constraint because the model can still use unlabelled features if they sufficiently reduce the loss). Another example of a knowledge constraint is \textbf{visibility}, which forces the neural network architecture itself to align with structured prior knowledge, such as a Gene Ontology database where each network node represents one ontology \cite{yu_visible_2018} (a hard constraint because the model is restricted to have a pre-defined architecture).

\subsection{Modularity}

Sufficiently complex models may require more than just simplifications to their input and connectivity, but may require a new architecture altogether.
We use the term \textbf{modularity} to describe architectures that decompose the total information contained within a big, complex model into smaller, simpler elements. These modules represent abstractions of the data that break up the whole learning process into distinct \textit{parts}, allowing the analyst to trace the relationship(s) between modules (being abstract concepts) as they would for features (being concrete measurements). In practice, modules are arranged to align with domain knowledge or human reasoning. For example, in computer vision, the appearance of an object and the motion of an object can be recognized by two separate modules--each with its own unique interpretable representations--that are combined together to make a final prediction. As another example, Li et al. proposed a neural network architecture for image detection that includes a module to learn visible prototypes of objects \cite{li_deep_2017}. When applied to hand-written digit prediction, the model representations can be inspected directly to visualize what a typical instance of each digit looks like \cite{li_deep_2017}.

\begin{figure}[H]
\centering
\scalebox{1}{
\includegraphics[width=(.8\textwidth)]{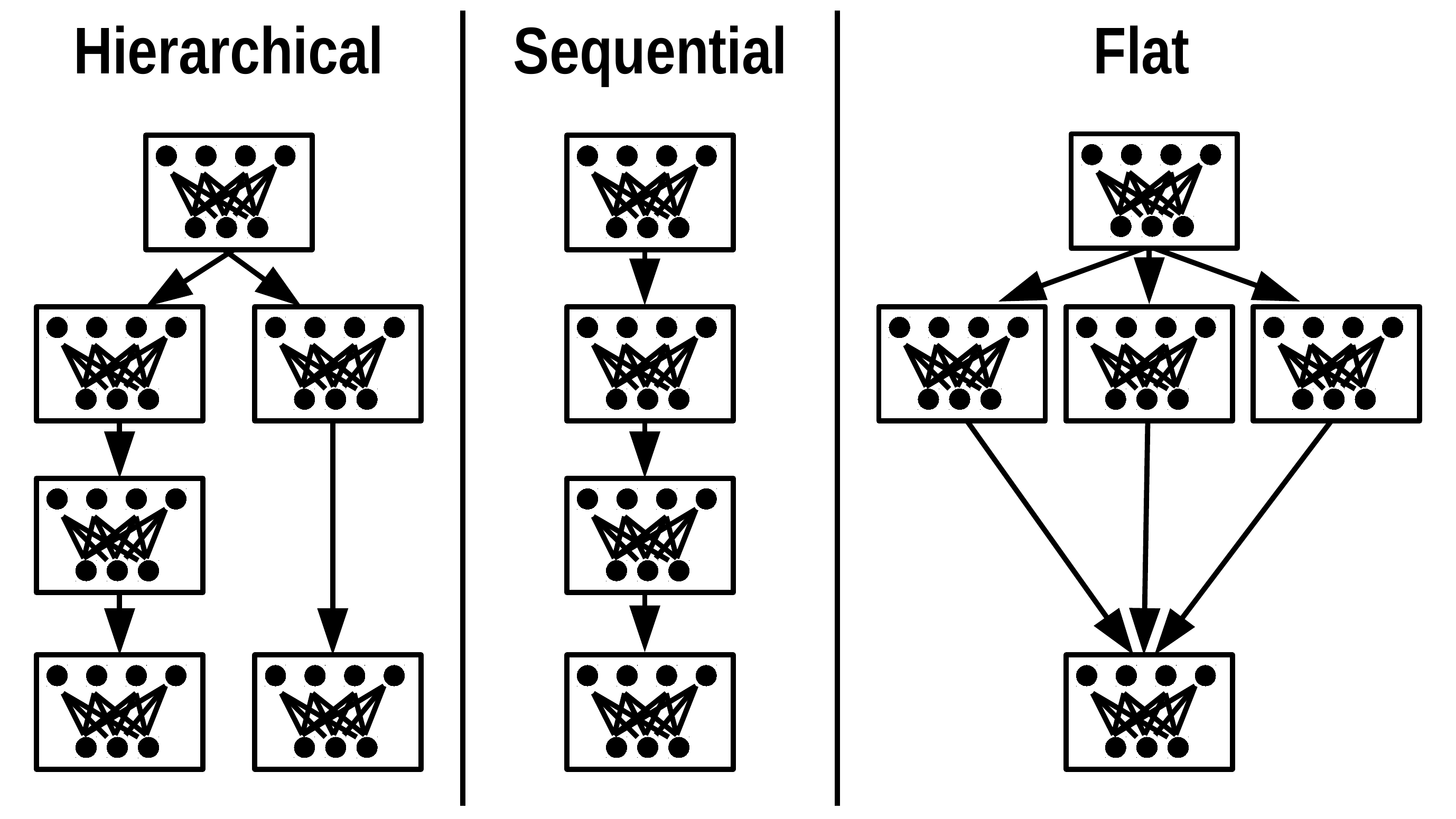}}
\caption{Types of modularity. In hierarchical and sequential arrangements, some modules are conditional upon other modules. In flat arrangements, each module is independent.}
\label{fig:module}
\end{figure}

Like constraints, modularity can be hard or soft, depending on whether it is introduced by a fixed architecture or by a regularization term. We can also classify modularity based on the arrangement of the modules. As depicted in Figure~\ref{fig:module}, the arrangement could be \textbf{hierarchical} where some parts build on top of previous parts, \textbf{sequential} where each part builds on top of all previous parts, or \textbf{flat} where each part is independent. Notably, these arrangements are all \textit{directed acyclic graphs} (DAGs), and so modularity shares some resemblance to causal inference methods that use DAGs to specify a prior causal structure within the data \cite{pearl_causal_1995}. Indeed, known causal structures can help inform module arrangement.

Hierarchy can improve model interpretation because humans naturally organize concepts hierarchically. 
In deep learning, hierarchical arrangements are common in multi-task models, where a single input is used to predict multiple outputs. For example, a single DNA input can be passed to multiple modules, where each module is responsible for predicting the binding affinity of one unique transcription factor \cite{eraslan_deep_2019}.
A \textbf{sequence} is a special type of hierarchy that can give a complex task a clear order. For example, a deep image recognition network may first perform object detection, then object feature extraction, and then finally classification. Ideally, both the detected objects and their extracted features are interpretable (c.f., OpenFaceAI \cite{amos_openface_2016}). Of note, \textbf{boosting} is a sequential model that entails training a subsequent model on the residual of the previous model over 1 or more iterations. In this case, the first model captures most of the input-output relationship, while the second model captures some remaining part of the input-output relationship, and so on. These models, which one can think of as a sequence of modules, facilitate interpretations because they imply a rank-order (similar to principal components). For example, if features A and B are important in the first model, while features C and D are important in the second model, then we could say that A and B are most important, followed by C and D \cite{greenacre_variable_2018}.

Modularity is still useful when not organized hierarchically. \textbf{Flat} modules enable a model to have distinct ``blocks'' that correspond to some aspect of the data (hence \textbf{blocking}). Ideally, each block would contain a mutually exclusive chunk of information, enforced by the architecture or by regularization. Flat modules, being arranged in parallel, can be computed in parallel, making them potentially fast to train (c.f., transformers \cite{vaswani_attention_2017,clauwaert_novel_2020}). One example of blocking is disentanglement, where the neural network latent space is encouraged to learn complete concepts (a soft modularity because the latent space may not fully disentangle concepts). Another example is ensemble 
(a hard modularity because the use of ensembles is a fixed architecture choice).

One compelling example of blocking that may be relevant to bioinformatics is the neural interaction transparency (NIT) architecture \cite{tsang_neural_2018}. NIT uses a custom regularization term to re-organize a fully-connected neural network into separate fully-connected blocks. Each block is fed by a limited number of features, and regularized so that features only enter a block when their total joint contribution is more predictive than the sum of their individual contributions (a soft modularity because the blocking depends on regularization). Consequently, the incoming connections to a block define a readily interpretable interaction set. Compared with conventional methods, NIT can detect any non-linear interaction, not just multiplicative interactions \cite{tsang_neural_2018}.

\section{Transparent Model Design Workflow}

Given the number of transparent model design concepts available for use, how does one go about designing a transparent model? This is a difficult question with no single answer, but we find it helpful to draw an analogy to \textit{retrosynthesis}, a popular approach to planning organic chemical synthesis. Retrosynthesis begins by considering the target molecule, then decomposing that molecule into readily available starting materials. This is roughly how we approach transparent model design too. We propose \textit{model retrosynthesis}, which begins by considering the target \textit{prediction task}, then decomposing that task into readily available starting materials. In the case of neural networks, the starting materials are pre-defined layers (e.g., a fully-connected, convolutional, or recurrent layer), as well as the transparent model design concepts from above.

\begin{figure}[H]
\centering
\scalebox{1}{
\includegraphics[width=(\textwidth)]{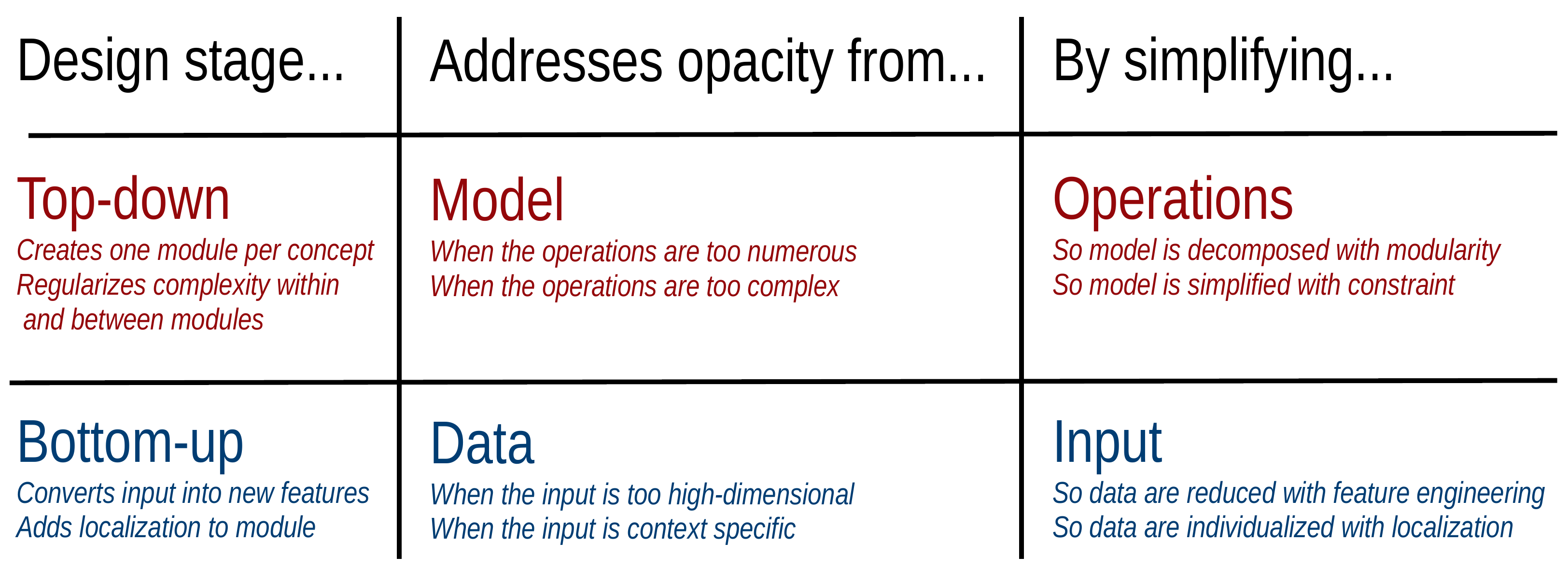}}
\caption{Two stages of transparent model design. The top-down stage addresses opacity arising from the model operations. The bottom-up stage addresses opacity arising from the data input. By simplifying both the model operations and the data input, the behaviour of a neural network can align more closely with the nature of human-scale reasoning.}
\label{fig:stages}
\end{figure}

We apply model retrosynthesis in 2 stages, as shown in Figure~\ref{fig:stages}. The first, which we call the \textit{top-down stage}, is optional, and exists to break down a big, complex prediction task into multiple smaller, simpler prediction sub-tasks. The second, which we call the \textit{bottom-up stage}, exists to make each task (or sub-task) transparently interpretable. In this way, we can start with the prediction task (i.e., the output), then build up modules and layers that will encode the relevant features (i.e., the input).
Each stage addresses one source of model opacity:

\begin{itemize}
    \item \textbf{Top-down stage} resolves opacity arising from the model, notably from when the operations are too numerous or too complex. During this stage, the prediction task is decomposed into sub-tasks. Each sub-task is assigned a module representing abstract but human-understandable concepts. The modules are related to one another by a computational graph that defines the relationship between the modules (e.g., hierarchical vs. flat) and the operations that relate them (e.g., addition, subtraction, multiplication, or complex non-linear). Operations within and between modules may be further limited using constraints.
    
    \item \textbf{Bottom-up stage} resolves opacity arising from the data, notably from when the input is too high-dimensional or inappropriately treated \textit{en masse}. During this stage, the input to each module is encoded into human-interpretable representations, typically through feature engineering. When appropriate, localization can be further added to the module. If the resulting module is still too complex, the top-down stage can be iterated over again to decompose the module further into even simpler sub-modules.
\end{itemize}

\begin{figure}[H]
\centering
\scalebox{1}{
\includegraphics[width=(.7\textwidth)]{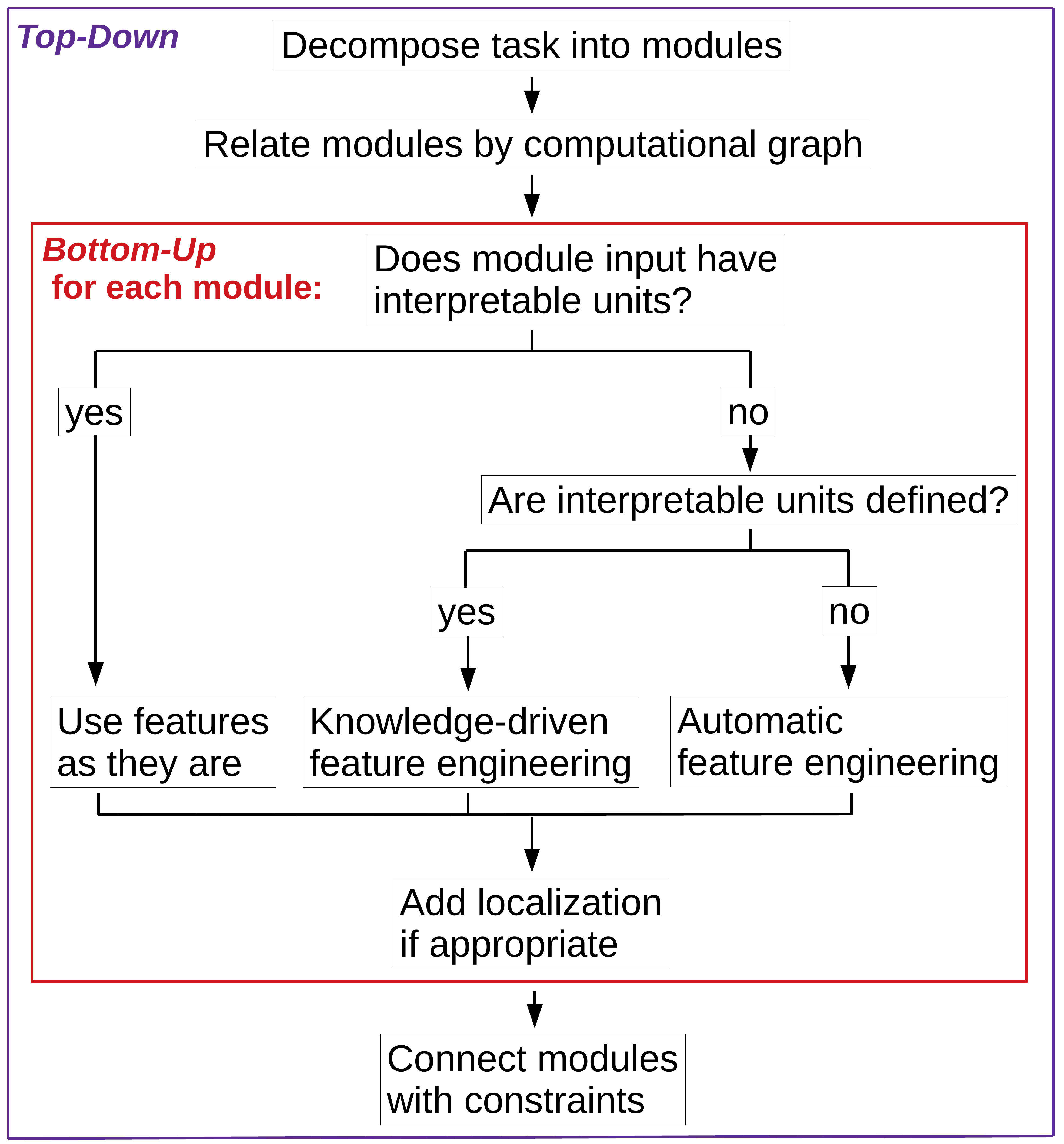}}
\caption{A flow diagram to guide model retrosynthesis.}
\label{fig:design-workflow}
\end{figure}

Figure~\ref{fig:design-workflow} presents a flow diagram that can help guide an analyst through the two stages of transparent model design. The \textit{top-down stage} begins by asking what modules, as human-understandable processes, might contribute to the overall prediction task. Then, it asks how these modules, as processes, relate to one another. Answering these questions involves making a \textit{computational graph} that describes the arrangement of the modules (e.g., hierarchical vs. flat), but also the mathematical operations that define the arrangements (e.g., addition, subtraction, multiplication, or complex non-linear). This is where it becomes helpful to draw upon known causal structures, when available, because they may imply a meaningful computational graph that naturally aligns with domain knowledge and human reasoning. When the relevant causal structures are unknown or incomplete, the analyst may have to guess about the computational graph.

The computational graph defines a set of modules that require a technical implementation during the \textit{bottom-up stage}. We consider 4 questions that help frame the implementation. First ask, ``What do we want to interpret?'' Second ask, ``How can we represent what we want to interpret within the model?'' Third ask, ``How can we use existing neural network architectures to achieve the desired representation?'' Fourth ask, ``How can we further constrain the network so that the desired representation is properly learned?'' In answering these questions, we start by thinking about the features. Are the features already interpretable? If so, you can use the features as they are and directly interpret the weights that connect them to the next module (akin to how one would typically interpret the coefficients of a linear model). If not, you will need to engineer the features. Is an interpretable abstraction already defined? If so, knowledge-driven feature engineering is a good choice. Otherwise, additional neural network layers are needed to perform the abstraction automatically.
Fully connected layers can help learn aggregates of features that work together, especially when used in conjunction with sparsity constraints \cite{le_deep_2020}. CNN layers are good for learning image-like structures which, in the bioinformatics domain, include DNA and RNA \cite{koo_representation_2019}. 
A description of other interpretable layers is beyond the scope of this article, and also a future direction in this field of research. Finally, localization is added if appropriate.

After the \textit{bottom-up stage}, it may be helpful to return to the \textit{top-down stage} to introduce constraints across the network, either within or between modules, to encourage the network to learn more useful representations from the data.

Note that while we present the workflow in a straightforward manner, model retrosynthesis is not always straightforward, and may require re-iterating upon various ideas before finalizing the design. The distinction between the top-down and bottom-up stages may become blurred in practice. Our intention is not to prescribe a rigid approach to model design, but rather to suggest a general framework that might help the reader hone their own creative design practice.

\subsection{A Case Study in Transparent Model Design}

As a simple but concrete example, let us align the above workflow with a previously published neural network architecture called DeepCoDA \cite{quinn_deepcoda_2020}, a model used to predict clinical phenotypes from microbiome compositions. We begin at the \textit{top-down stage} by considering the motivations behind the work. The first motivation is to transform the compositions into log-ratio biomarkers. The second motivation is to assign personalized importance scores to the biomarkers. Accordingly, the authors decompose the overall prediction task into two modules: (1) a biomarker discovery module, used to extract log-ratio biomarkers from the data, and (2) a localization module, used to personalize the association between the biomarkers and the clinical phenotypes. The relationship between these modules is sequential. The first module extracts biomarkers which get passed to the second module to learn local linear fits.
This concludes the modularization phase of the top-down stage, resulting in a basic computational graph (without forks or colliders) that we can use during the bottom-up stage. Note that modularization did not require us to write any code; rather, the purpose of modularization is to establish a road map to help guide the implementation later.

Next, we consider the \textit{bottom-up stage}. We begin with the biomarker discovery module. According to the authors, the input does not have interpretable units. Moreover, the desired abstractions are not already known (ruling out knowledge-driven feature engineering). Thus, a layer is added to represent the desired interpretable units via automatic feature engineering. In this case, the motivation is to learn log-ratio biomarkers, so a custom layer is used to learn the ratios directly (a novelty of the work). Next, we move on to the localization module. The input to this module are the ratios, which, according to the authors, are already interpretable units, and so no further abstraction is required. Instead, an attention-like layer is added to introduce personalization with local linear fits (adapted from \cite{alvarez-melis_towards_2018}). Finally, we return again to the \textit{top-down stage}, where constraints are added within the modules to encourage the learned representations to take on values consistent with the desired interpretation.

Note that one could instead collapse the two modules into a single module that performs both biomarker discovery and localization. For many problems, there will exist multiple equally valid ways to arrange the modules. We advise you use the module arrangement that most helps you to understand the problem at hand, and to explain your solution to others.

\section{Reporting Transparent Design}

When conducting machine learning as part of an academic research project, the researcher may wish to communicate their transparent model design to a wider audience. It can be difficult to decide on what and how much detail to include when writing a scientific study. Reporting guidelines can help improve the quality and accessibility of scientific papers \cite{montenegromontero_transparency_2019}.

\begin{figure}[H]
\centering
\scalebox{1}{
\includegraphics[width=(\textwidth)]{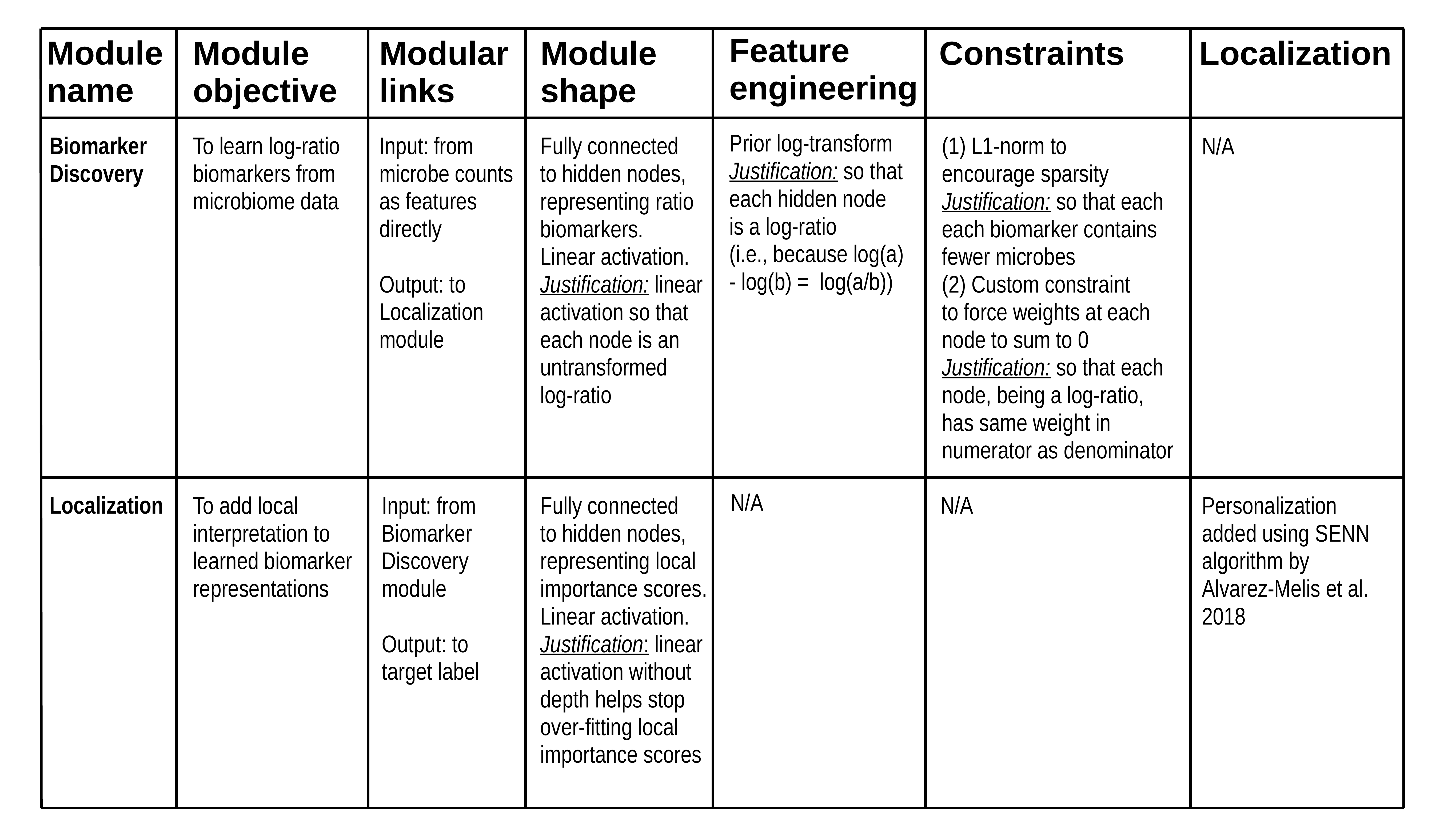}}
\caption{A template for transparent model design reporting, using the DeepCoDA case study as an example. The rows refer to modules, while the columns refer to aspects of the modules.}
\label{fig:template}
\end{figure}

Above, we described a conceptual framework for transparent model design. This framework can also help structure scientific reporting. For example, researchers can describe their new model with regard to the 4 taxonomic categories: (1) feature engineering, (2) localization, (3) constraint, and (4) modularity. Transparent model design reporting should go beyond the ``What?'' of design choices, to also cover the ``Why?''. Researchers should discuss how their choices help address both model opacity and data opacity, and thus allow the model to achieve human interpretability. Unlike other reporting guidelines, which typically provide a checklist, we think that, given the vast possibility space of design choices and their implications, a template may have broader utility. In our template, the rows describe distinct modules (or layers), while the columns describe key properties of the modules (or layers). Relevant properties include the shape and connectivity of the modules, as well as any feature engineering, localization, or constraint used within them. The report should not only describe what techniques were used, but also provide a brief justification for their use. In this way, the template can trace the model from input to output, while giving the reader insights into how each module serves the interpretability objective.
As an example, Figure~\ref{fig:template} presents a template using the DeepCoDA case study from above. 

Note that our proposed template only covers transparent model design reporting and does not, for example, cover data collection, data pre-processing, model training, model tuning, or model verification (all equally important aspects of machine learning workflows). We recommend that researchers also consider other relevant reporting guidelines, especially when conducting biomedical research, such as Prediction model Risk Of Bias ASsessment Tool (PROBAST) \cite{wolff_probast_2019} and Transparent reporting of a multivariable prediction model for individual prognosis or diagnosis (TRIPOD) \cite{moons_transparent_2015} guidelines. While not specific to machine learning, they include many elements that are useful for evaluating the completeness of a machine learning publication \cite{nagendran_artificial_2020}. 
Other machine learning and AI-specific reporting guidelines are under active development and will likely have an important role in the near future \cite{consort-ai_and_spirit-ai_steering_group_reporting_2019}.

\section{Summary}

As deep learning becomes a mainstay of scientific research, we expect that scientists will slowly move away from using neural networks for the sole purpose of prediction, and will instead aim to use neural networks for the purpose of scientific discovery. Such a change in the motivation behind deep learning may require a change in how deep learning models are designed, namely from the design of \textit{opaque, uninterpretable models} towards the design of \textit{transparent, interpretable models}. Although transparent model design is inherently a creative process, there are established design concepts that can help improve the interpretability of neural networks, and thus facilitate scientific discovery. Above, we proposed a taxonomy of transparent model design concepts, along with the workflow for putting design concepts into practice, that we hope will enable others to use explainable AI (XAI) and scientific XAI. Good luck!

\section*{Declarations}

\subsection*{Author Contributions}

TPQ and VL formulated the modularization procedure and the transparent model design workflow. TPQ formulated the taxonomy and wrote the manuscript. SG, SV, and VL provided technical expertise and helped revise the manuscript.

\subsection*{Competing Interests}

No authors have competing interests.

\subsection*{Acknowledgements}

Not applicable.

\bibliographystyle{unsrt}
\bibliography{references,extra}

\end{document}